\def\year{2019}\relax
\documentclass[letterpaper]{article} 
\usepackage{aaai19}  
\usepackage{times}  
\usepackage{helvet}  
\usepackage{courier}  
\usepackage{url}  
\usepackage{graphicx}  
\usepackage{caption}
\usepackage{booktabs}
\usepackage{enumitem}
\usepackage{paralist}
\usepackage{xspace}
\usepackage{multirow}
\usepackage{bm}
\usepackage{amsmath}
\usepackage{amsfonts}
\usepackage{amsthm}
\usepackage{amssymb}
\usepackage{todonotes}
\usepackage{lipsum}
\usepackage{subcaption}
\DeclareMathOperator{\E}{\mathbb{E}}

\begin{document}
%
\newcommand{\gram}[1]{\textbf{\textcolor{red}{gn: #1}}}
\newcommand{\xiang}[1]{\textbf{\textcolor{blue}{xk: #1}}}

\title{An Adversarial Approach to High-Quality, \\ Sentiment-Controlled Neural Dialogue Generation}
\author{Xiang Kong, Bohan Li, Graham Neubig,  Eduard Hovy, Yiming Yang\\
  Language Technologies Institute\\
  Carnegie Mellon University\\
  \{\tt xiangk, bohanl1, gneubig, hovy, yiming\}@cs.cmu.edu
}
\maketitle
\begin{abstract}
In this work, we propose a method for neural dialogue response generation that allows not only generating semantically reasonable responses according to the dialogue history, but also explicitly controlling the \emph{sentiment} of the response via sentiment labels. Our proposed model is based on the paradigm of conditional adversarial learning; the training of a sentiment-controlled dialogue generator is assisted by an adversarial \emph{discriminator} which assesses the fluency and feasibility of the response generating from the dialogue history and a given sentiment label. Because of the  flexibility of our framework, the generator could be a standard sequence-to-sequence (SEQ2SEQ) model or a more complicated one such as a conditional variational autoencoder-based SEQ2SEQ model. Experimental results using automatic and human evaluation both demonstrate that our proposed framework is able to generate both semantically reasonable and sentiment-controlled dialogue responses.
\end{abstract}

\section{Introduction}
\label{sec:intro}

Sentiment is a fundamental part of the human communication, and reflecting sentiment in human-computer interfaces is a key to making them engaging and interesting to use.
This is certainly true for dialog systems and there has been a large body of literature that attempts to equip dialogue systems with the ability to understand and express the sentiment~\cite{polzin2000emotion,skowron2011good,partala2004effects,prendinger2005empathic,hasegawa-EtAl:2013:ACL2013,zhou2017emotional,shen2017conditional}. However, these methods are either based on templates or rules which require extensive hand engineering.

End-to-end neural dialogue generation~\cite{shang2015neural,vinyals2015neural,serban2016building} is now a popular research topic because of the ease and flexibility of creating systems within this paradigm. While early works~\cite{shang2015neural,vinyals2015neural} just employ simple sequence-to-sequence (SEQ2SEQ) models similar to those used in machine translation~\cite{cho2014learning,sutskever2014sequence}, a number of papers have aimed to further improve the quality and diversity of dialogue responses in manners specific to dialog systems~\cite{li2015diversity,li2016deep,gu2016incorporating,xing2016topic,li2017adversarial,serban2017hierarchical,zhao2017learning}.

Furthermore, There have also been a few attempts~\cite{skowron2011good,hasegawa-EtAl:2013:ACL2013,shen2017conditional,zhou2017emotional} to incorporate sentiment information into data-driven end-to-end dialog systems, but each has their own shortcomings. For example, \citeauthor{hasegawa-EtAl:2013:ACL2013}~\shortcite{hasegawa-EtAl:2013:ACL2013} propose a method to train individual systems for each kind of emotion, which will cause the system to suffer from data sparsity and high computational cost.
In addition, \citeauthor{shen2017conditional}~\shortcite{shen2017conditional} incorporate latent variables expressing emotion into the dialogue system but do not provide an explicit way to control the sentiment of these responses.

Recently, \citeauthor{acl2018zhou}~\shortcite{acl2018zhou}  collect a large corpus of tweets from Twitter with emojis in the response, and assume that these emojis could reflect the sentiment of the response. Furthermore, they train a conditional variational autoencoder (CVAE)-based neural dialogue system which is capable of controlling the sentiment of the generated response explicitly. In this work, we investigate the application of another powerful model, i.e., generative adversarial networks (GANs) to this problem.

In this paper, we propose a conditional generative adversarial network (CGAN)-based framework for sentiment-controlled dialogue generation. In this framework, the desiderata of fluency and controlability are explicitly enforced by creating a model with two subcomponents: a generator and a discriminator. 
The generator is in charge of generating sentimental responses given a dialogue history and a sentiment label, while the adversarial discriminator enforces sentimental response quality by trying to determine whether the item (\textit{dialogue history, sentiment label, dialogue response}) comes from the real data distribution. By training the generator to fool the discriminator, our system can simultaneously improve the quality of dialogue responses and generate responses with different sentiments depending on the sentiment label. 
\section{Background}
\label{sec:background}
\subsection{Problem Setting}
Our task is to train a dialogue system which is able to generate high-quality and sentiment-controlled responses. Given the dialogue history $\mathbf{W}_{h}=\{\mathbf{w}_{1}, \ldots, \mathbf{w}_{i}, \ldots,  \mathbf{w}_{h_{N}}\}$, where $\mathbf{w}_{i}$ is the i-th token and $h_{N}$ is the length of the sequence, and a sentiment label $y$, the task is to generate a response $\mathbf{W}_{r}=\{\mathbf{w}_{1}, \ldots, \mathbf{w}_{i}, \ldots,  \mathbf{w}_{r_{N}}\}$ where $r_{N}$ is the length of the generated response.  We would like this response to be consistent with the sentiment label $y$ and semantically appropriate for the dialogue history $\mathbf{W}_{h}$.

\subsection{Encoder-Decoder for Dialog Generation}\label{sec:seq2seq}

Most neural models for dialogue generation are based on the encoder-decoder structure, a.k.a, sequence-to-sequence (SEQ2SEQ) framework~\cite{cho2014learning,sutskever2014sequence}, in which an encoder reads in the previous dialog history/context and encodes it a continuous vector representation $c$, which the decoder then uses to output the next dialogue utterance.

Specially, given the dialogue history $\mathbf{W}_{h}$, the hidden state of the encoder at time $t$, $h_{t}^{enc}$, is computed according to:
\begin{equation}\label{eq:1}
h_{t}^{enc}=\mathbf{RNN}_{enc}(h_{t-1}^{enc},\mathbf{w}_{t}),\,t=1,...,h_{N},
\end{equation}
where $\mathbf{RNN}_{enc}$ is the encoder RNN. Finally, we obtain the vector representation of $\mathbf{W}_{h}$, i.e.,  $c=h_{h_{N}}^{enc}$.

The decoder is another RNN which is capable of generating a response $\mathbf{W}_{r}$ given the context vector $c$. The hidden state of the decoder at the time step $t$ is calculated by: 
\begin{equation}\label{eq:2}
h_{t}^{dec}=\mathbf{RNN}_{dec}(h_{t-1}^{dec},c,\mathbf{w}_{t-1})
\end{equation}
where $\mathbf{RNN}_{dec}$ is the decoder RNN.

The probability $p_{t}^{dec}$ over the whole vocabulary at $t$-th time step is then calculated by a softmax function conditioned on the hidden state $h_{t}^{dec}$. 

By multiplying all probabilities of the gold word tokens at each time step, we can calculate the probability $p(\mathbf{W}_{r}|\mathbf{W}_{h})$ of the response sequence $\mathbf{W}_{r}$ given the dialogue history sequence $\mathbf{W}_{h}$.

\subsection{Generative Adversarial Nets}

The second important technology contributing to the proposed method is Generative Adversarial Networks (GANs; \citeauthor{goodfellow2014generative}~\shortcite{goodfellow2014generative}).

In the original GAN framework, there are two models: a generative model $G$, which is in charge of generating outputs (one example being the SEQ2SEQ model in the previous section), and a discriminative model $D$ that attempts to discriminate whether its input samples are real or generated outputs.
By training $G$ to create outputs that are able to fool $D$ into thinking that they are real, it is possible to generate samples that seem highly realistic, improving the quality of generation of images~\cite{salimans2016improved} and text~\cite{li2017adversarial,yu2017seqgan}.

However, one major problem in the GAN framework is that there is no mechanism to control attributes of generated items. Therefore, \citeauthor{mirza2014conditional}~\shortcite{mirza2014conditional} propose a condition adversarial nets (CGANs) in which both the generator and discriminator are conditioned on some extra information so that the generator can control the types of items being generated according to this extra information.

Back to the dialogue scenario, \citeauthor{li2017adversarial}~\shortcite{li2017adversarial} propose an adversarial dialogue generation model, in which the generative model $G$ is a standard SEQ2SEQ model~\cite{sutskever2014sequence} which could generate a response $\mathbf{W}_{r}$ given the dialogue history $\mathbf{W}_{h}$ according to Eq.\ref{eq:1} and Eq.\ref{eq:2}.  The discriminative model $D$ is a binary classifier which takes the dialogue history $\mathbf{W}_{h}$ and a dialogue response  ${\mathbf{W}_{r}}$ as an input and outputs a label $D(\mathbf{W}_{h},\mathbf{W}_{r})$  indicating whether the dialogue response ${\mathbf{W}_{r}}$ is generated from machines or human beings. 

In more detail, its objective is to maximize the expected reward, i.e., $D(\mathbf{W}_{h},\mathbf{W}_{r})$ of generated responses :
\begin{equation}\label{eq:5}
J=\E_{\mathbf{W}_{r}\sim G_{\theta}(\cdot|\mathbf{W}_{h})}[D(\mathbf{W}_{h},\mathbf{W}_{r})]
\end{equation}

\subsection{Variational Autoencoders}
Another popular deep generative model recently is the framework of variational autoencoders (VAEs).
VAEs have been successfully applied to many text generation tasks~\cite{bowman2015generating,serban2017hierarchical,zhou2017multi,hu2017toward}. Specifically, in~\citeauthor{serban2017hierarchical}~\shortcite{serban2017hierarchical}, the dialogue generation model has been augmented by introducing a latent variable $z$ at the decoder. \citeauthor{shen2017conditional}~\shortcite{shen2017conditional} and \citeauthor{acl2018zhou}~\shortcite{acl2018zhou} present a CVAEs-based framework for dialogue generation in which the response $\mathbf{W}_{r}$ is generated from  a stochastic latent variable $z$ and and the context vector $c$. Mathematically, a CVAE-based dialogue generation system maximizes a variational lower bound on the conditional likelihood of $\mathbf{W}_{r}$ given the latent variable $z$ and the context vector $c$.  

\section{Approaches}\label{sec:app}
In this section, we build upon the SEQ2SEQ-based dialogue generation model with the CVAE and CGAN techniques introduced in the previous section.

\subsection{Sentiment-Context SEQ2SEQ}
In order to explicitly control the sentiment of the generated response, we slightly change the structure of the standard SEQ2SEQ model~\cite{sutskever2014sequence}. Specifically, after obtaining the context vector $c$ of the dialogue history $\mathbf{W}_{h}$ from the encoder by Eq.\ref{eq:1}, the concatenation of a sentiment vector $s$ and $c$ is fed into the decoder to generate a response and this vector is called ``sentiment context'', $s_{c}$. 

To generate the sentiment vector $s$, similarly to word embedding, we first map the sentiment label $y$ to a vector $y_{e}$, and this vector will be fed into a fully-connected neural network to output the sentiment vector $s$.

\begin{figure}[t]
	\centering
	\includegraphics[width=0.49\textwidth]{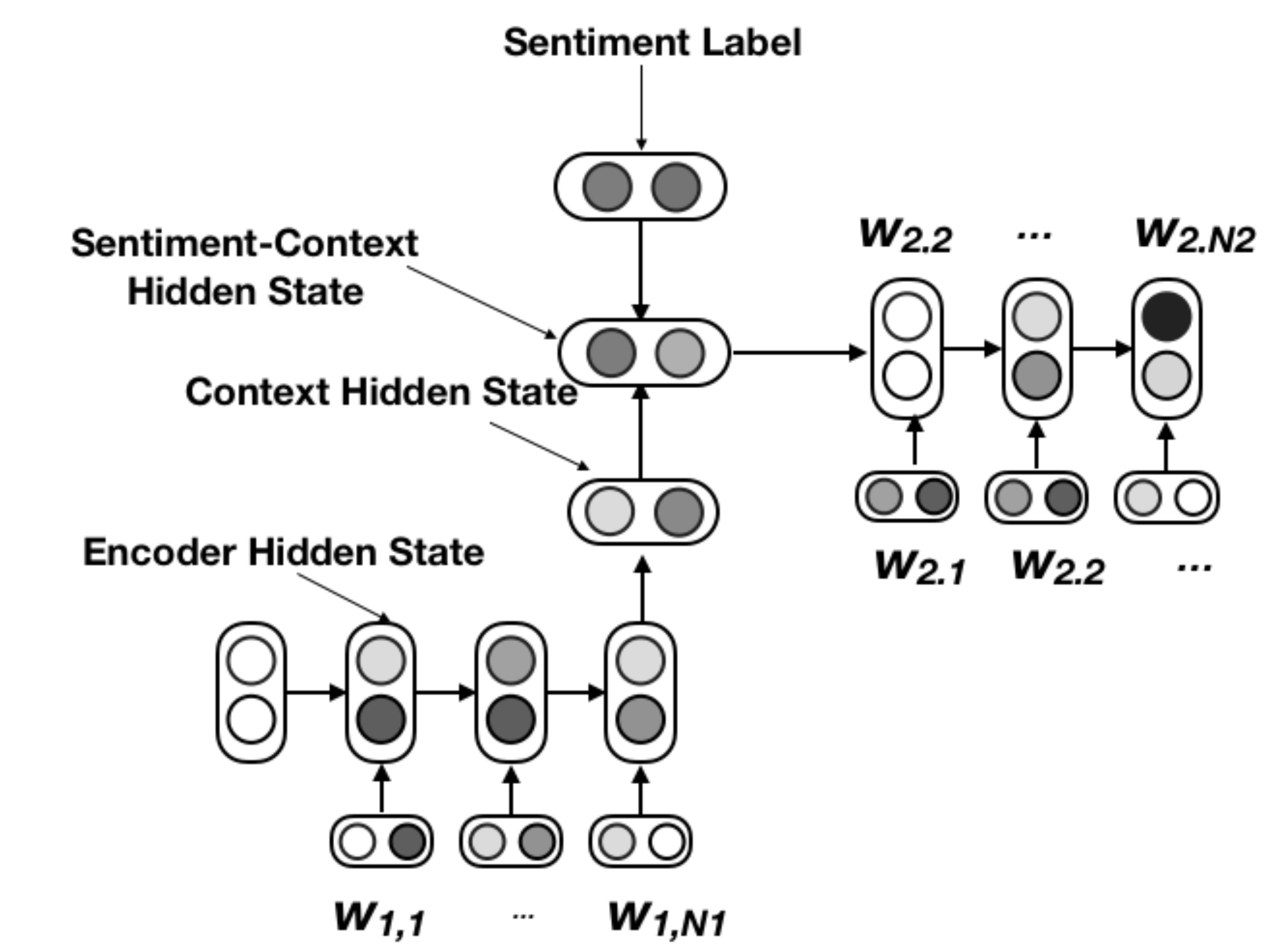}
	\caption{The computational graph of the sentiment-context SEQ2SEQ architecture. The dialogue history is encoded into a dense vector $c$ via an encoder RNN, then the concatenation of the context vector $c$ and the sentiment vector $s$ computed from the sentiment label $y$ is fed into a decoder RNN to generate tokens in the response.}
	\label{fig:base}
\end{figure}

The computation graph of the Sentiment-Context SEQ2SEQ model is shown in Figure~\ref{fig:base}.

\subsection{Conditional Variational Autoencoders (CVAEs) SEQ2SEQ}\label{sec:cvae}
We follow the model structure described in \citeauthor{sohn2015learning}~\shortcite{sohn2015learning} and \citeauthor{acl2018zhou}~\shortcite{acl2018zhou} to build the CVAE-SEQ2SEQ model.

Mathematically, the objective of CVAE-SEQ2SEQ is to maximize the lower bound probability of the response given the sentiment context vector, i.e.,
\begin{equation}
    p(\mathbf{W}_{r}|s_{c})= \int p(\mathbf{W}_{r}|z,s_{c})p(z|s_{c})dz
\end{equation}
where $z$ is the latent variable, $s_{c}$ is the sentiment context vector mentioned before. Based on the assumption that the latent variable follows a multivariate Gaussian distribution
with a diagonal covariance matrix, the lower bound of $\log p(\mathbf{W}_{r}|s_{c})$ is:
\begin{equation}
    \begin{split}
        \mathcal{L}_{\textrm{CVAE}}=&~\mathbb{E}_{q(z|\mathbf{W}_{r},s_{c})}(\log p(\mathbf{W}_{r}|z,s_{c})\\
        &-\textrm{KL}(q(z|\mathbf{W}_{r},s_{c})||p(z|s_{c})
    \end{split}\label{eq:elbo}
\end{equation}
where $p(\mathbf{W}_{r}|z,s_{c})$ is modeled by another decoder which is different from the decoder in the seq2seq model, $q(z|\mathbf{W}_{r},s_{c})$ is described by a recognition network and $p(z|s_{c})$ is modeled by a prior network, both of which are MLP-based neural networks..

In more detail, for the encoder RNN, CVAE-SEQ2SEQ uses the same setting as the SEQ2SEQ model to encode the dialogue history $\mathbf{W}_{h}$ into a context vector. The decoder RNN, however, is different because it now takes the concatenation of the sentiment context vector and the sampled stochastic latent variable as input to generate a response. At training time, the latent variable sample $z$ is drawn from an approximate posterior network and used for optimizing the variational lower-bound given by Eq. \ref{eq:elbo}. At test time, the latent variable sample $z$ is drawn from a prior network for decoding, which has no knowledge of the ground-truth response. Furthermore, the bag-of-word loss~\cite{zhao2017learning} has been added in the above objective function. Therefore, the final objective function for the CVAE-SEQ2SEQ is:
\begin{equation}
    \mathcal{L} = \mathcal{L}_{\textrm{CVAE}} + \mathcal{L}_{bow}
\end{equation}


\subsection{Conditional Generative Adversarial Net SEQ2SEQ}
\label{sec:cgan}
\begin{figure}[t]
	\centering
	\includegraphics[width=0.49\textwidth]{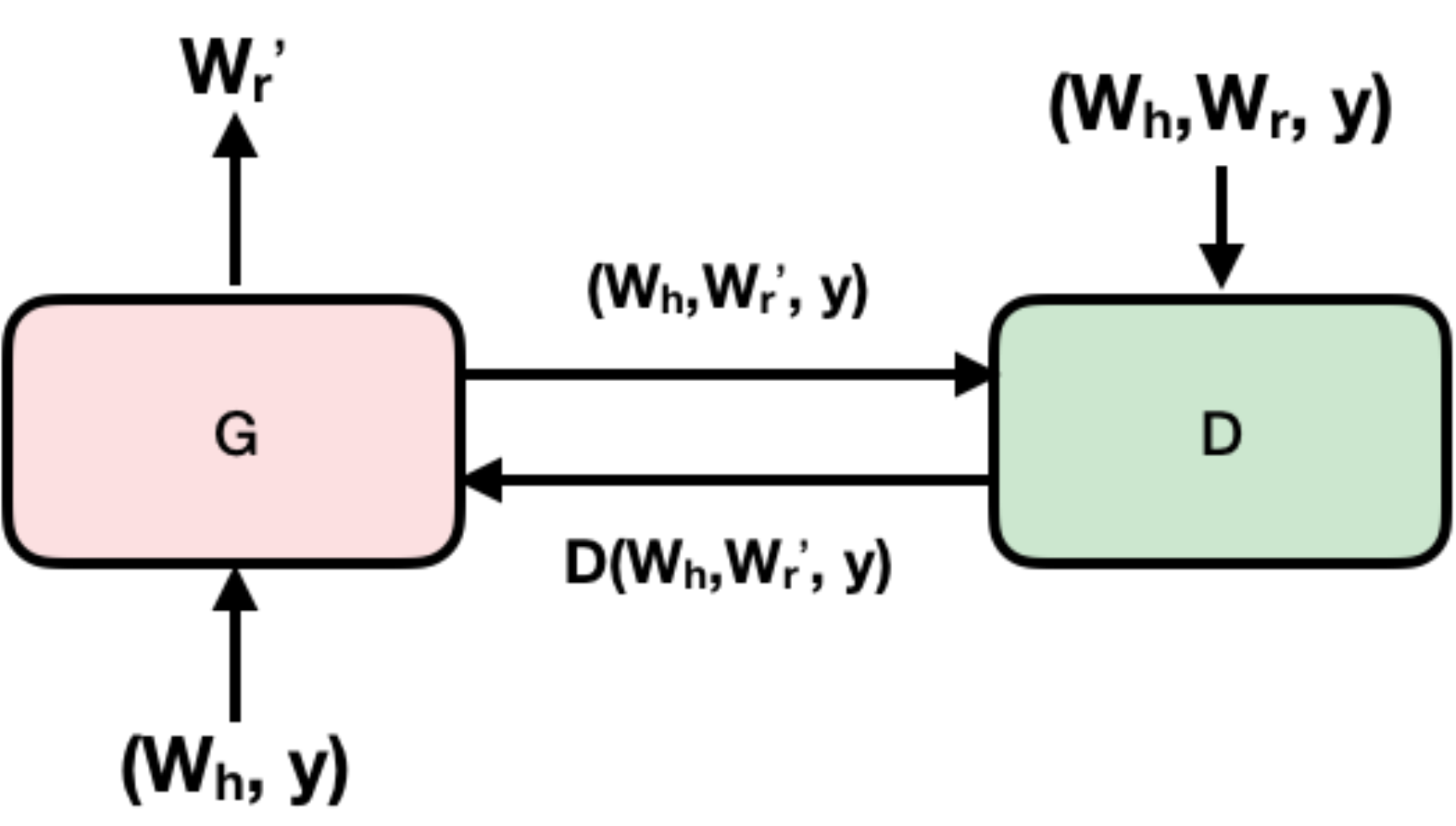}
	\caption{The computational graph of the conditional generative adversarial networks-based SEQ2SEQ architecture. ``G'' denotes the generator  and ``D'' refers to the discriminator. $W_{r}^{'}$ is the response being generated from the generator.}
	\label{fig:cgan}
\end{figure}

\begin{figure}
    \centering
    \includegraphics[width=0.49\textwidth]{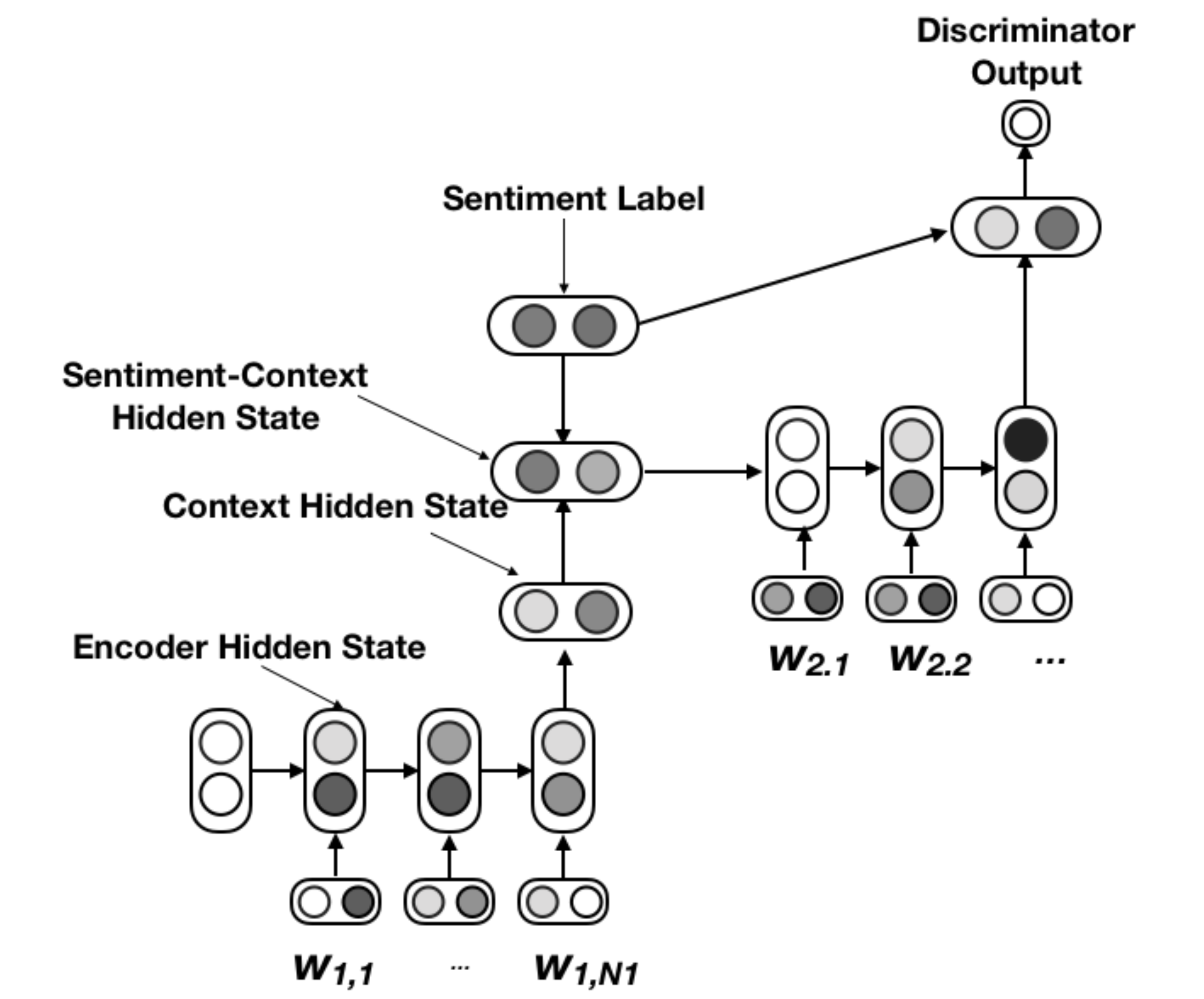}
    \caption{The discriminator structure in the CGAN-SEQ2SEQ model. After obtaining the final vector from the second encoder, the concatenation of this vector and the sentiment context vector will be fed into an MLP layer for the final output. }
    \label{fig:dis}
\end{figure}

Adversarial training methods have been successfully applied to neural dialogue generation~\cite{li2017adversarial} to improve the quality of generated responses. However, in their model, the property of the response such as the sentiment, could not be controlled explicitly. Therefore, we propose a conditional generative adversarial network-based dialogue system named CGAN-SEQ2SEQ which is able to improve the quality of the response and control its sentiment at the same time. The model is shown in Figure~\ref{fig:cgan}. 

Our proposed CGAN-SEQ2SEQ consists of two components, i.e., a conditional generator $G$ and a conditional discriminator $D$.

\noindent \textbf{Generator $G$}. We adopt the original sentiment-context SEQ2SEQ as the $G$ which could generate a response $\mathbf{W}_{r}$ given the dialogue history $\mathbf{W}_{h}$ and a sentiment label $y$. 
The goal of the $G$ is to produce high-quality and sentiment-controlled responses as similar as those being generated from human beings so as to fool the discriminator $D$. 

\noindent\textbf{Discriminator $D$}. The discriminator $G$ in our framework is to identify whether the input response is generated from human beings or machines given the dialogue history $\mathbf{W}_{h}$ and the sentiment label $y$. 
Specifically, the discriminator consists of two encoders. The first encoder is similar to that in the SEQ2SEQ model which is able to encode the input dialogue history to a representation vector which will be concatenated with the sentiment vector to compose the sentiment context vector. For the second encoder, the initialize state will be set as the sentiment context vector allowing the decoder to condition on the $\mathbf{W}_{h}$ and $y$, then it encodes the response sequence to a representation vector. Finally, the concatenation of this vector and sentiment context vector will be fed into a fully-connected neural network-based binary classifier to compute the final result. The reason why we utilize the sentiment context vector again is to let the $D$ pay more attention to the sentiment information. The computation graph of the $D$ is illustrated in the Figure~\ref{fig:dis}.

\noindent\textbf{A Game with Two Players}. Following the training process mentioned in the \citeauthor{li2017adversarial}~\shortcite{li2017adversarial}, we first pre-train a generator without the discriminator, then freeze the parameters of the pre-trained generator to pre-train the discriminator.
During pre-training of the discriminator, responses generated from the pre-trained generator and human beings are regarded as negative and positive samples respectively. Finally, 
the generator $G$ and the discriminator  $D$ play a two-player game. In this game, the generator first generates a response $\mathbf{W}_{r}^{'}$ given the dialogue history $\mathbf{W}_{h}$ and the sentiment label $y$, then the discriminator  provides $D(\mathbf{W}_{r}^{'}, \mathbf{W}_{h}, y)$ back to the generator and use triples ($\mathbf{W}_{h}$, $\mathbf{W}_{r}^{'}$,  $y$) and ($\mathbf{W}_{h}$, $\mathbf{W}_{r}$,  $y$) to train itself.  The generator will be optimized according to the $D(\mathbf{W}_{r}^{'}, \mathbf{W}_{h}, y)$ obtained from discriminator.


\noindent\textbf{Policy Gradient Training}. Similarly to \citeauthor{li2017adversarial}~\shortcite{li2017adversarial}, the generator in our proposed framework is a probabilistic transformation from the dialogue history to the dialogue response, both in discrete space. Therefore, we also employ the REINFORCE algorithm~\cite{williams1992simple} to optimize it.

The objective of the generator is to maximize the expected reward of generated responses: 
\begin{equation}\label{eq:cgan}
J=\E_{\mathbf{W}_{r}^{'}\sim G_{\theta}(\cdot|\mathbf{W}_{h},y)}[D(\mathbf{W}_{h},\mathbf{W}_{r}^{'},y)]
\end{equation}.

Note that $D(\mathbf{W}_{h},\mathbf{W}_{r}^{'},y)$ can be regarded as the probability of the response $\mathbf{W}_{r}^{'}$ being generated from human beings given the $\mathbf{W}_{h}$ and $y$. The gradient with respect to the $\theta$ in Eq.\ref{eq:cgan} could be approximately computed by the likelihood ratio trick~\cite{williams1992simple}:
\begin{equation}
\begin{split}
   {\triangledown_{\theta}} \approx  [D&(\mathbf{W}_{h},\mathbf{W}_{r}^{'}, y) - b(\mathbf{W}_{h},\mathbf{W}_{r}^{'}, y)]\\
   &{\triangledown_{\theta}}\sum_{t}\log p(w_{t}|\mathbf{W}_{h}, y, W_{1:t-1}) 
\end{split}\label{eq:cgan-gradient}
\end{equation}
where $b(\mathbf{W}_{h},\mathbf{W}_{r}^{'}, y)]$ is the baseline value of the expected reward which could reduce the variance of the estimate while keeping it unbiased~\cite{ng1999policy}. 

Intuitively, when the generated response is more likely to fool the $D$, the larger reward the $G$ will get, and thus parameters will be updated with a larger step.

One advantage of the CGAN-SEQ2SEQ over CVAE-SEQ2SEQ is that it will not change the structure of the SEQ2SEQ model. During response  generation, the discriminator could be removed and the SEQ2SEQ model remains the same.

We found training to be unstable if we just use Eq.\ref{eq:cgan-gradient} to optimize the $G$. Therefore, we also employ the teacher forcing procedure~\cite{li2017adversarial} to assist the training so that the generator has access to the golden response.

\subsection{CGAN-CVAE SEQ2SEQ}

From the Figure~\ref{fig:cgan}, it is easy to find that the generator could also be the CVAE-SEQ2SEQ. Therefore, we propose a CGAN-CVAE  SEQ2SEQ model, in which the generator is the CVAE-SEQ2SEQ model and the discriminator stays the same as that in the CGAN SEQ2SEQ model. Intuitively, from the reinforcement learning perspective, the discriminator is regarded as the reward provider and for a high-quality generated response, it will assign a higher reward back to the generator. 

In order to stabilize the training process, besides adding the teaching forcing method mentioned in the previous section, we also add the original CVAE objective to the Eq.~\ref{eq:cgan} to create a hybrid objective function~\cite{acl2018zhou}, i.e., 
\begin{equation}
\begin{split}
    \mathcal{L}^{'}=&\E_{\mathbf{W}_{r}^{'}\sim G_{\theta}(\cdot|\mathbf{W}_{h},y)}[D(\mathbf{W}_{h},\mathbf{W}_{r}^{'},y)] + \mathcal{L}
\end{split}
\end{equation}
\section{Experimental Results}
\subsection{Dataset}
To evaluate our proposed framework, we use the large corpus of tweets with emojis collected and used in \citeauthor{acl2018zhou}~\shortcite{acl2018zhou}. To simplify the task, we classify all emojis into two clusters, i.e., positive and negative. As a result, there are approximately 374$\mathbf{K}$, 21$\mathbf{K}$ and 21$\mathbf{K}$ tweets in train, dev and test sets respectively. The ratio of the ratio of positive to negative samples is around 3:1.

\subsection{Evaluation Metrics}
\noindent\textbf{Perplexity:}
Perplexity is a common metric used in many natural language tasks and connected to the likelihood of the gold response given a dialogue generation model. Although the diversity of responses generated by the dialogue system is very important, the system should nonetheless assign a relatively high likelihood to the ground truth response. 

\noindent\textbf{Sentiment Accuracy:} Because the goal of our task is to control the sentiment of the response given a sentiment label and a dialogue history, whether the generated response correctly reflects the sentiment is very important. Therefore, we build a sentiment classifier on the training set and evaluate the generated responses by how often the classifier-predicted label represents the specified sentiment $y$. 

\noindent\textbf{Human Evaluation:} Because automatic metrics are sub-optimal for evaluating performance of dialogue generation systems~\cite{liu2016not}, we ask three judges  to evaluate 30 random items, each of which consists of a dialogue history, a gold response, and a generated response. Judges are expected to evaluate in two settings.
\begin{compactitem} 
\item In one setting, the goal is to evaluate the quality of dialogue responses from different models.
We use a 1-5 scale where $5$ means that the response and the dialogue history is highly relevant semantically and syntactically, and $1$ means they are irrelevant.
\item In the other setting, judges are asked to label the sentiment of the given responses as positive or negative.
In this case only the generated responses are provided to the judges. 
\end{compactitem} 
Note that these two experiments are conducted separately and the items are different in order to avoid bias.

\subsection{Implementation Details}
\noindent\textbf{Sentiment Classifier} Our sentiment classifier is 1-layer bidirectional RNN-GRU encoder with 128 hidden units in each direction.
This is fed into an MLP classifier to predict the final sentiment class. 

We employ a standard SEQ2SEQ~\cite{sutskever2014sequence} model with attention~\cite{luong2015effective} to build the sentiment-context SEQ2SEQ model. The encoder is a 1-layer bidirectional GRU with hidden size 128 in each direction, and the dimension of the sentiment vector is 12.  The decoder is a 1-layer GRU of size $128*2+12=268$. The Adam optimizer with a 1e-3 learning rate and gradients clipped to 5 is employed to train this model. 

Following the experiment settings in \citeauthor{acl2018zhou}~\shortcite{acl2018zhou}, we incorporate a response encoder, a recognition network and a prior network into the above SEQ2SEQ model to build a CVAE-SEQ2SEQ model. The response encoder is another 1-layer bidirectional GRU of size 128 in each direction. The mean and log variance of latent variable $z$ is obtained from the recognition and the prior network, both of which are two fully-connected networks, then latent variables are sampled via the reparameterization trick~\cite{kingma2013auto}. During generation without golden responses, the latent variable sampled from the prior network will be directly fed into the decoder.

\subsection{Main Results}
\noindent\textbf{Capacity for Sentiment Control:} The sentiment control capacity of each model is evaluated by the sentiment accuracy metric. As shown in Table~\ref{tab:results}, the CGAN-CVAE SEQ2SEQ model outperforms all the other models in sentiment accuracy, indicating that, combining CGANs and CVAEs together, the generator could control the sentiment of the response more effectively than the respective baselines. Although the sentiment accuracy of the CGAN-SEQ2SEQ is better than the SEQ2SEQ model, it can not control the sentiment of the response as well as the CVAE-SEQ2SEQ model. We suspect that this is because during REINFORCE training the generator can only access the generated sentences, which will be noise to deteriorate the generator if they are of low quality. We have found that the responses from the pre-trained generator are indeed generic and do not control the sentiment information well. However, the CVAE-SEQ2SEQ model can utilize the golden response at every training step.

\noindent\textbf{Response Quality:} We employ Perplexity (PPL), which is shown in Table \ref{tab:results}, as a proxy to evaluate the response quality. Compared with other models, the CGAN-CVAE SEQ2SEQ model achieves the lowest PPL score, which means that its likelihood of generating the golden response is highest. Similarly to the sentiment accuracy, the PPL of the CGAN-SEQ2SEQ is higher than that of the CVAE-SEQ2SEQ and we attribute this to the same reason mentioned above.

\subsection{Human Evaluation}
The human evaluation result is shown in Table~\ref{tab:human-results}. With respect to both content quality and sentiment accuracy, the CGAN-CVAE has better accuracy than other models. This demonstrates that our proposed CGAN-CVAE could not only generate high-quality dialogue response but effectively control the sentiment of dialogue responses as well, which is consistent with the automatic evaluation results. The overall performance of the CVAE model is also better than that of the CGAN model. 

\subsection{Case Study}
In order to show the differences between the performances of these models more concretely, we show some examples in Table~\ref{tab:case_study}. We can clearly see that the responses generated from the CGAN-CVAE SEQ2SEQ and CVAE-SEQ2SEQ models are more distinctive given different sentiment labels and topics related to the dialogue context. For CGAN-SEQ2SEQ, the sentiment of the response is relatively consistent with the sentiment label but compared with CVAE-SEQ2SEQ, the diversity of responses is relatively low. As for the SEQ2SEQ model, it seems that it only remembers some sentimental words and the responses are quite dull and generic. 

\begin{table}[!ht]
\centering
 \begin{tabular}{||c c c||} 
 \hline
  & Perplexity & Sentiment Acc (\%) \\ [0.5ex] 
 \hline\hline
 SEQ2SEQ & 157.5  & 55.6 \\ 
 \hline
 CVAE & 81.83 & 75.6 \\
 \hline 
 CGAN & 120.3  & 64.4 \\
 \hline
 CGAN-CVAE & \textbf{69.54} & \textbf{78.8} \\
 \hline
\end{tabular}
 \caption{Evaluation of various dialogue systems with perplexity and sentiment accuracy. }\label{tab:results}
\end{table}

\begin{table}[!ht]
	\centering
	\begin{tabular}{||c c c||} 
 \hline
  & Quality & Sen-Acc(\%) \\
 \hline\hline
 SEQ2SEQ & 2.1  & 54.4 \\ 
 \hline
 CVAE & 3.6  &  73.3\\
 \hline
 CGAN & 2.9 & 66.7 \\
 \hline
 CGAN-CVAE & \textbf{3.9} & \textbf{78.9} \\
\hline
\end{tabular}
 \caption{Dialogue response quality and sentiment accuracy (Sen-Acc) of different dialogue systems based on human evaluation. }\label{tab:human-results}
\end{table}

\begin{table*}[!ht]
\centering
 \begin{tabular}{p{3cm}|p{6cm}|p{6cm} }
 \hline
 Context & \multicolumn{2}{|l}{goldlink is dope live one of my favorite shows i 've been to}  \\ 
 \hline
 Sentiment & Positive & Negative \\
 \hline 
 SEQ2SEQ & i 'm so happy for you & i 'm so sad \\
 \hline 
 CGAN &i 'm gonna be there  & i 'm not sure i 'm gonna be able to find it \\
 \hline 
 CVAE & i like the song  &i feel like i was gonna cry   \\
 \hline 
 CVAE-CGAN & omg i love it  & that 's the worst \\
 \hline
  \hline
 Context & \multicolumn{2}{|l}{and i never got $\langle dgt\rangle$
 lol}  \\ 
 \hline
 Sentiment & Positive & Negative \\
 \hline 
 SEQ2SEQ &  i 'm so excited & i 'm so sorry  \\
 \hline 
 CGAN & he 's so cute  &  i 'm not sure i 'm not going to be a fan of $\langle dgt\rangle$ \\
 \hline 
 CVAE & i love to hear that ! i 'm so happy to hear this   & well , didn 't realize you had to get the wrong name  \\
  \hline 
 CVAE-CGAN &  lmao i ’ m looking for it  &i 'm sorry for you\\
 \hline
   \hline
 Context & \multicolumn{2}{|l}{always got ya my dude no matter what ! lets bowl}  \\ 
 \hline
 Sentiment & Positive & Negative \\
 \hline 
 SEQ2SEQ & i'm so mad & i 'm not sure if it 's a good idea\\
 \hline 
 CGAN &i 'm glad you 're enjoying it  &  i 'm not sure if you 're joking\\
 \hline 
 CVAE & we are doing a great job ! & wow . i hate you  \\
 \hline 
 CVAE-CGAN &  yes i love you guys , but it is a good time   &i mean i hate my bestfriend\\
 \hline
\end{tabular} 
 \caption{Response samples from different dialogue models given different sentiment labels. }\label{tab:case_study}
\end{table*}
\section{Related Work}


The sentiment is crucial to human-human communication, and thus for machines to communicate smoothly with humans, it is necessary for machines to generate utterances with sentiment.

\citeauthor{skowron2011good}~\shortcite{skowron2011good} propose that affective profiles in a dialogue system are strongly correlated with the emotional changes experienced by participants. \citeauthor{partala2004effects}~\shortcite{partala2004effects} show that positive affective intervention could be especially useful to enhance users' problem solving performance. \citeauthor{prendinger2005empathic}~\shortcite{prendinger2005empathic} indicate that a computer agent with empathic feedback could support people preparing for an interview. \citeauthor{polzin2000emotion}~\shortcite{polzin2000emotion} describe how a dialogue system adjusts its interaction according to the emotion of users.  
 
While these works are valuable proofs-of-concept, they generally either focus on small-scale corpora or using rule-based templates to generate responses, which make them difficult to extend to large-scale open-domain dialogue generation.

Some recent works have tried to incorporate sentiment in large-scale conversation generation.
\citeauthor{hasegawa-EtAl:2013:ACL2013}~\shortcite{hasegawa-EtAl:2013:ACL2013} study how utterances affect the emotion of other speakers, and try to predict the emotion of the user to generate a response which reflects the specific emotion using the statistical machine translation framework~\cite{ritter2011data}.
Within the framework of neural dialogue systems, pioneering work by \citeauthor{shen2017conditional}~\shortcite{shen2017conditional} incorporate sentiment into the variational hierarchical encoder-decoder model~\cite{serban2017hierarchical}. However, their model is trying to improve the quality of the response instead of controlling the sentiment of the response.  
In parallel, \citeauthor{zhou2017emotional}~\shortcite{zhou2017emotional} tried to model sentiment in a dialogue conversation system through three mechanisms: emotion category embedding, internal emotion memory, and external memory. The internal memory models the change of the internal emotion state of the decoder, and therefore encodes how much an emotion has already been expressed. The external memory decides whether to choose an emotional or generic (non-emotional) word at a given step during decoding. 

\section{Conclusion}

In this paper, we propose an intuitive training objective for neural dialogue generation, which is to control the sentiment of the generated response explicitly. This objective is implemented through a conditional adversarial training paradigm, in which the generator is trained to generate sentiment-controlled responses via sentiment labels assisted by a discriminator. Furthermore, the generator in our system could be the standard SEQ2SEQ or CVAEs-based SEQ2SEQ models. Our system adopts a policy gradient algorithm to deal with the optimization challenge posed by discrete generator outputs.  Experiments clearly demonstrate the effectiveness of such an adversarial training objective in successfully controlling the sentiment of the response and improving the content quality. 

Future directions include validating our approach on more fine-grained sentiment-based data and improved combination with other advanced techniques in reinforcement learning and adversarial learning such as reward shaping, etc.

\newpage
\bibliography{bib.bib}
\bibliographystyle{aaai.bst}

\end{document}